# Short-term Prediction of Household Electricity Consumption Using Customized LSTM and GRU Models

Saad Emshagin, Wayes Koroni Halim, Rasha Kashef
Electrical, Computer, and Biomedical Engineering
Toronto Metropolian Univeristy
{Semshagi, wayes.halim, rkashef}@ryerson.ca

*Abstract*—With the evolution of power systems as it is becoming more intelligent and interactive system while increasing in flexibility with a larger penetration of renewable energy sources, demand prediction on a short-term resolution will inevitably become more and more crucial in designing and managing the future grid, especially when it comes to an individual household level. Projecting the demand for electricity for a single energy user, as opposed to the aggregated power consumption of residential load on a wide scale, is difficult because of a considerable number of volatile and uncertain factors. This paper proposes a customized GRU (Gated Recurrent Unit) and Long Short-Term Memory (LSTM) architecture to address this challenging problem. LSTM and GRU are comparatively newer and among the most well-adopted deep learning approaches. The electricity consumption datasets were obtained from individual household smart meters. The comparison shows that the LSTM model performs better for home-level forecasting than alternative prediction techniques-GRU in this case. To compare the NN-based models with contrast to the conventional statistical technique-based model, ARIMA based model was also developed and benchmarked with LSTM and GRU model outcomes in this study to show the performance of the proposed model on the collected time series data.

## I. Introduction

As we move toward a sustainable power system, we can see that renewable energy sources, the increasing number of electric vehicles or EVs, and the cyclical requirement of power on distribution grids will unsurprisingly contribute to a power system that is highly complex and unpredictable. There is no doubt that developing a stable system has never been more difficult than it is today, thanks to the variable nature of renewable energy. This is due to the complicated interplay between utilities and customers. The role of load forecasting has become increasingly important as the modern electric system has evolved, developed and played a major part in modernizing the system. According to the prediction period, load forecasting can be classified into three types: i) short-term, ii) medium-term, and iii) long-term [1] forecasting. Generally, short-term prediction considers a time window from an hour to a week. Medium-term predictions can last from a week to a year, whereas long-term predictions are applicable when the time window is more than a year. Short-term prediction is used to decide which power plants to use when to schedule power systems, and how to forecast the energy market [1, 2]. The long- and medium-term forecasts are essential to the design of future sites and determining the source of fuel for power plants [1] and [2]. It is, therefore, very important to project the electricity requirement of residential customers in the short term to the extent that this can significantly facilitate the long-term operation of the power system. It is possible to reduce peak load with smarter technology for demand response or adaptable energy storage systems, given the load prediction has high accuracy and when peak load is forecasted effectively. When utility companies can get precise load projections for individual consumers, they can apply this data to target the best customer groups with the strongest probability of participating in disaster recovery programs when the power goes out in case of power shortages. It will be more likely for the utilities to reduce peak demand if they implement a Demand Response program based on incentives, helping them achieve a more stable energy reserve and better cost management by reducing their demand [3] [4].

As a result of the stochastic nature of consumer behavior and climate conditions, it is extremely difficult to forecast short-term electricity consumption when it comes to an individual household unit [5]. A high degree of accuracy can be achieved when managing short-term loads on an aggregated basis compared to managing short-term loads on an individual basis [3]. Globally, smart meter infrastructure (SMI) development is paving the way for conventional power systems to be replaced by smart grids. Moreover, deploying many such systems has opened up potential solutions for short-term load prediction for unit customers [5]. A smart meter provides suppliers and consumers with near-real-time information regarding the cost and the amount of energy consumed. Homes receive more precise bills, useful information about their energy use patterns, and hour-based pricing when energy use data is recorded with a higher resolution [6]. There are several methods available for forecasting electrical load. In general, these methods can be divided into two categories.

These are mainly - i) conventional methods and ii) artificial intelligence methods [6]. The majority of traditional approaches use statistical techniques [2], [7], [8], [9]. Some examples of these methods are Multiple Linear Regression (MLR) [2], [7], [8], [9], Auto-Regressive Integrated Moving Average (ARIMA) [2], [7], [8], [9], and Exponential Smoothing (ES) [2], [8], [9], etc. A notable limitation of the aforementioned methods is that they are not always effective in forecasting short-term load demands due to the nonlinear nature of time-series load data. Moreover, general forecasting methods, such as MLR or ARIMA, are often employed for forecasting aggregated system loads [6]. The first study implied statistical models to predict short-term household loads using time series analysis [10][11]. Later on, as the development of artificial intelligence progressed, various machine learning methods and deep learning-based techniques were built and applied to anticipate household energy demand. Some remarkable examples would be Artificial Neural Networks (ANNs) [2], [7], [8], [9], Recurrent Neural Network (RNN) based models like Long Short-Term Memory (LSTMs) [2], [3], [4], [10], and Gated Recurrent Unit (GRU) [36], etc. Apart from these - Radial Basis Functional Networks (RBFNs) [2], [8], [9], and hybrid methods [1], [2], [11] are also some examples. As a technology, deep learning has recently become one of the most popular research topics in a number of research extends around the world. In contrast to shallow learning, deep learning usually consists of stacking neural networks over multiple layers. It also uses stochastic optimization to perform machine-learning tasks. Various levels of complexity can be achieved by varying the number of layers to improve the ability to learn and perform a task [12]. Hochreiter and Schmidhuber [13] first introduced Long Short-Term Memory (LSTM), a type of Recurrent Neural Network (RNN), which was able to attract a great deal of interest in sequence learning research. It has been shown that LSTM networks can be applied to translate natural language, put captions for images, and recognize speech [14] and other applications [15]-[18]. While they handle a lot of data classification, they have limited applications for regression tasks.

This paper proposes a framework for forecasting short-term load using modified LSTM and GRU units in a deep learning network. A single set of distinct time series is used to test both models. These time series represent household data from smart meters. Following this, we describe a framework for forecasting residential loads based on LSTMs and GRUs, intending to show how they can record historical consumption data for an individual meter-based household and make a competent forecast in our case. Nevertheless, we considered how the NN-based models, LSTM and GRU in our case, performed in contrast to statistical-based models, such as ARIMA. An ARIMA-based model was developed and compared with the evaluation outcome of the two models, as mentioned earlier within the additional scope of this paper, to learn how appropriate this model (ARIMA-based) would be when dealing with time series data. Last but not least, we demonstrate that by aggregating individual forecasts, we can generally get better forecasts than direct forecasts on aggregated loads. Using five prediction-based metrics, we evaluated the performance of the LSTM, GRU, and ARIMA model: Root Mean Square Error (RMSE), Mean Squared Error (MSE), R-squared Score, Mean Absolute Error (MAE), and Mean Absolute Percentage Error (MAPE).

A summary of the rest of the paper follows - Section II discusses an overview of related work on load prediction. Section III discusses the proposed forecasting framework and models. Section IV discusses the methods used in this approach and summarizes the outcome. Section V concludes the work and suggests possible directions for further research

## II. LITERATURE REVIEW

Providing an accurate energy load forecast is crucial for making informed decisions in the energy sector. Load forecasting is used in many industries and by researchers to determine the energy needed to maintain a demand-supply equilibrium. Thus, researchers have developed numerous methods for predicting future electricity demand. In Cao et al. [19], an ARIMA-based model was employed to forecast intraday loads. Using the average demand on the same days in each time range, their similar day method predicts the load for the specific day based on past days with high climate similarity in history. On ordinary days, the ARIMA method performs better, while on unusual days, the "similar day" method developed by the author is more effective [19]. [20] and [21] used a Neural Network model with a radial basis function to forecast short-term loads. Yun et al. [20] combined an RBF neural network with an Adaptive Neural Fuzzy Inference System (ANFIS) to fine-tune the model to consider the real-time cost of electricity. The grid method is combined with backpropagation (BP) and RBF neural networks by Li et al. [21] for short-term day-ahead forecasting. Like [19], the grid method groups load profiles based on location, nature, and size rather than by days based on similar meteorological measurements. Here, a training dataset for each group was used that applied a backpropagation network, along with a radial basis function, to project a load forecast for the next consecutive day. Qingle et al. [22] also presented an NN-based load forecast for brief periods. To predict the load value at the next time step, only the current and previous time step values were used by them. According to Zhang et al. [23], extreme learning machines (ELMs) were applied to learn the energy load and forecast the total load for the Australian national energy market. As well as using the ELM learning model for self-adaptive learning, the proposed methodology utilized ensemble learning to minimize any event of unstable prediction. There has also been some successful applications of the K-Nearest Neighbor (KNN) algorithm to predict electricity load [24], [25]. Ghofrani et al. developed a dedicated input selection scheme for hybrid forecasting frameworks using a Bayesian Neural network [26]. On average, this approach achieved a MAPE score of 0.419%. All these techniques, however, aim to learn and forecast load at a grid level of the energy system. Effective load prediction approach for electricity users is becoming increasingly critical to aid smart grid applications. In their study, Zhang et al. [27] selected a predictive load forecasting model, aiming for load clustering using smart meter data and a decision tree. A similar methodology was adopted by Quilumba et al. [28] that utilizes smart meter data for clustering and neural networks to forecast load for the system during the hours of a day. The data of individual households have been clustered and labeled by Stephen et al. [29]. The individual households were then regarded as label sequences, which were further fitted into their model. Then, a day ahead label was sampled, and the mean value for the cluster at each instant was utilized to forecast the consumption for the following day. Thus, it is possible to reduce forecasting errors by clustering different customers based on their daily usage profiles. Nevertheless, error in forecasting aggregated usage is likely to occur as multi-user diversity offsets individual customer error predictions.

Two examples of load forecasting for individual users are Chaouch's work [30] and Ghofrani et al. [31] in the existing literature. Chaouch et al. [30] proposed a useful forecasting approach with time-series data and reported the median absolute errors, but validation using MAPE was not mentioned. Thus, it cannot be used for benchmarking for trial comparisons. MAPE ranging from 18% to 30% was calculated in Ghofrani et al.'s work [31]. Deep learning-based methods have recently been developed in load forecasting. Based on Ryu et al. [32], deep neural networks can improve load forecast accuracy for industrial customers. Because industrial power consumption trends are more frequently regular than residential consumption, the MAPE on average is much more accurate than that in [31], which results in an average MAPE of 8.85%. For single-meter residential load forecasting, Mocanu et al. [33] applied a deep learning method with factored Boltzmann machine, with significant performance improvements compared to shallow ANN and SVM [3]. Also, Marino et al. [34] formulated LSTM and demonstrated comparable results to [33] in their attempt to forecast load. As a result, comparing the two pioneering works with others is difficult since their effectiveness was only evaluated based on the RMSE (Root Mean Square Error) metric instead of MAPE, which is a more common metric. Moreover, [34] failed to provide a justifiable explanation of why LSTM was chosen. Additionally, no aggregated impact can be determined for user-wise energy prediction when relating all the different approaches in [30], [31], [33], and [34].

III. RECURRENT NEURAL NETWORKS (RNN)

By connecting the output and hidden layers, recurrent neural networks incorporate sequential data from the input layer in a more advanced manner than conventional neural networks. A memory unit in RNN helps it predict new values using previous output results or data from a hidden layer. It is a helpful feature to improve the prediction [6] considerably. RNNs can learn correlation over time; however, with a longer delay between the input information and the prediction point, the performance is highly impacted by a long-term dependency problem. A loss of learning ability occurs when historical data becomes untraceable because of a phenomenon called gradient exploding or gradient vanishing [3], [6], [35]. This problem can be solved by a complex recurrent unit, such as LSTMs (Long Short-Term Memory) and GRUs (Gated Recurrent Units) [36].

A. LSTM

The LSTM network is presented as an improved version of the RNN [13], which can be seen in Fig. 1.

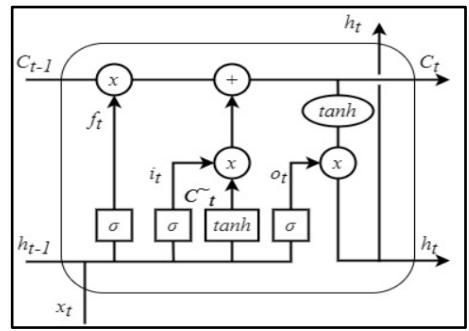

Fig. 1. LSTM Network Architecture [13]

There are three main gates in LSTM networks: input gate, output gate, and forget gate. At time t, the input data is the output of the same network at time t-1, and it is the output of the previous network at time t-1 [6]. Data from the former layers of the first layer is considered input. Input gates remember new and previous steps, while forget gates delete insignificant stuff from memory cells. Features will be transferred over the LSTM network during the training phase. The output gate might apply useful data from the memory cell. As a result, deep LSTM networks are excellent at figuring out the trend of complicated objects. The necessary equations for an LSTM network are [35]:

$$i_t = \sigma(W_i \cdot [h_{(t-1)}, x_t] + b_i) \quad (1)$$

$$f_t = \sigma(W_f \cdot [h_{(t-1)}, x_t] + b_f) \quad (2)$$

$$o_t = \sigma(W_o[h_{(t-1)}, x_t] + b_o) \quad (3)$$

$$\tilde{C}_t = \tanh(W_c[h_{(t-1)}, x_t] + b_c) \quad (4)$$

$$C_t = f_t \otimes C_{(t-1)} + i_t \otimes \tilde{C}_t \quad (5)$$

$$h_t = o_t \otimes \tanh(C_{(t)}) \quad (6)$$

Three sigmoid functions are presented in equations (1) through (3), where $x_t$ represents the input at a time step t. The gate parameters for weight are the W's, and the gate parameter for biases are the b's for input, forget, and output gates. The sigmoid function can be expressed as σ(x)=1/(1+e(-x)). The input gate is $f_t$, the forget gate is $i_t$, and the output gate is $o_t$. Equation (5) states that the tanh layer can make a new candidate value $\tilde{C}_t$ for t and add it to the cell state. Hidden states of LSTM units can be divided into two categories: "slow" states, $C_t$, and "fast" states $h_t$ [6]. A new slow state $C_t$ is generated by summing the multiplying of the forget gate $f_t$ by the previous cell state $C_{(t-1)}$ and the multiplying of the input gate it by the new candidate value $\tilde{C}_t$ [6]. To update the $h_t$ state, the hyperbolic tangent function (tanh) is used [6]. A key advantage of LSTM units is their ability to accumulate activities chronologically. The derivatives of the error do not disappear quickly over time [6], [13]. This allows LSTMs to handle long sequences of tasks.

### B. GRU

Another technique is GRU (Gated Recurrent Unit), presented in the work of Cho et al. [37]. The model identifies the dependent factors at various time scales and deals with the issue of vanishing gradients. Compared to LSTM, GRU has a missing memory cell [38], so the structure is simpler, resulting in less computation power and faster training. GRU can be used to address the problem statement because of the missing memory cells. The GRU has two gates, one for resetting, R, and one for updating, Z. The equation for the reset gate can be represented as follows [36].

$$R = sig(WR.C_{t-1} + VR.x_t) \quad (7)$$

The update gate determines how to retain the previous memory, update the content, and carry it forward [36]. This gate is computed by equation (8)

$$Z = sig(W_z c_{t-1} + V_z x_t) \quad (8)$$

The new memory in the GRU cell is determined based on both the new input and the past hidden state, as represented by equation (9) [36]. This gate will completely cancel out the past hidden state if the past hidden state is no longer relevant.

$$M = tanh(W_M(c_{t-1} * R) + V_M x_t) \quad (9)$$

By combining the past hidden input and the new memory, with the influence of the update gate, the hidden state is generated [36]:

$$c_t = (c_{t-1}) * (1 - z) + Z * M \quad (10)$$

W stands for weight matrices, and V stands for parameter vector. Also, the input, current output, and previous output are denoted by $x_t$, $c_t$, and $c_{t-1}$, respectively. The structure of the GRU cell is illustrated in Fig 2.

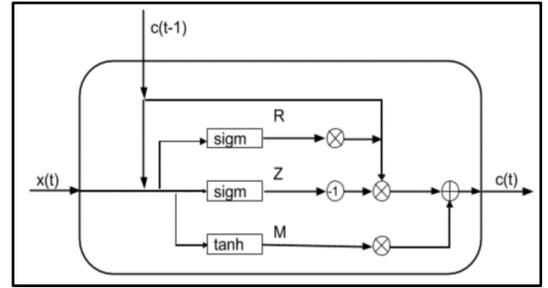

Fig. 2. Structure of GRU cell (Multiple interacting layers) [36]

The GRU unit is different from traditional RNNs in that it does not replace the content by adding a new value but keeps the content and adds new information to make it more accurate [36].

## IV. THE MODIFIED FORECASTING MODELS

In this paper, two prediction models were built to forecast power consumption at a household level using customized Long Short-Term Memory (LSTM) and Gated Recurrent Units GRU) based models, which are well suited for the time series dataset considered for this paper. Both models used the same overall structure, including the number of layers, units, and activation functions, to allow for performance comparison in later stages of the experiment. Each model comprises six layers where the first layer, the input layer, specifies a window size of 6 chosen for the experiment and the fact that these are single variant models. The 2[nd] and 4[th] layers use 140 units for the computation process. The models used a dropout rate of 40% in layer 3. The dropout rate is used to drop certain values, which may cause over or underfitting from the output of the preceding layers. The "Tanh" activation function is used in both models. Finally, the output goes through two final dense layers using the "Tanh" activation function.

```
LSTM_model = Sequential()
LSTM_model.add(InputLayer((6, 1)))
LSTM_model.add(LSTM(140,
activation='tanh',
return_sequences = True))
LSTM_model.add(Dropout(0.4))
LSTM_model.add(LSTM(140,
activation='tanh'))
LSTM_model.add(Dense(32, 'tanh'))
LSTM_model.add(Dense(1, 'tanh'))
LSTM_model.summary()
```

Fig. 3. The Customized Long Short-Term Memory (LSTM) model

```
GRU_model = Sequential()
GRU_model.add(InputLayer((6, 1)))
GRU_model.add(GRU(140,
activation='tanh',
return_sequences = True))
GRU_model.add(Dropout(0.4))
GRU_model.add(GRU(140,
activation='tanh'))
GRU_model.add(Dense(32, 'tanh'))
GRU_model.add(Dense(1, 'tanh'))
GRU_model.summary()
```

Fig. 4. The Customized Gated Recurrent Units GRU) model

Furthermore, the models are compiled with the mean Squared Error as a loss function, Adam optimizer with a learning rate of 0.0001, and Root Mean Squared Error as a scoring metric. Finally, the models are fitted using 100 epochs concerning a checkpoint used to commit the model with the least loss value to memory for later use. The batch size is set to 64, as shown in Fig.5.

```
checkpoint = ModelCheckpoint('LSTM_model/',
save_best_only=True)

LSTM_model.compile(loss=MeanSquaredError(),
optimizer=Adam(learning_rate=0.0001),
metrics= RootMeanSquaredError ())

LSTM_model.fit(x_train,y_train,
validation_data=(x_val,y_val),      epochs=100,
batch_size= 64, callbacks=[checkpoint])
```

Fig. 5. Loss Function and Optimizers

## V. EXPERIMENTAL ANALYSIS AND DESIGN

Due to the non-linearity nature of the dataset, certain prediction models may be considered for the experiment, namely Recurrent Neural networks-based techniques such as Long Short-Term Memory (LSTM) and Gated Recurrent Units GRU). Furthermore, the type of dataset and nature of the proposed forecasting model dictate most of the parameters utilized in building and designing the models. For instance, the "Tanh" activation function is better suited for time series predictions than ReLU or other known activation functions. Therefore it is utilized in building forecasting models. Most other parameters are selected based on long trials and errors. For instance, a window size of the last six hours to forecast the next one hour's data produced the best results compared to other higher and lower window size values previously considered. Table 1 summarizes various parameters used in the forecast models.

TABLE I.    LIST OF HYPERPARAMETERS USED IN THE FORECAST MODELS

| Parameter | Value |
| --- | --- |
| Activation function | Tanh |
| Number of epochs | 100 |
| Batch size | 64 |
| Loss | Mean Squared Error |
| Evaluation metric | Root Mean Squared Error |
| Optimizer and Learning rate | Adam, 0.0001 |
| Number of layers for each model | 6 |
| Window size | The previous 6 hours to forecast the next hour |

### A. Data sources and Preprocessing

This paper considers an actual household power consumption dataset for one of the authors. The data was obtained from the customer portal for Waterloo North Hydro. Hourly usage data was downloaded between August 10[th] and November 5[th], 2022, as shown in Fig. 6.

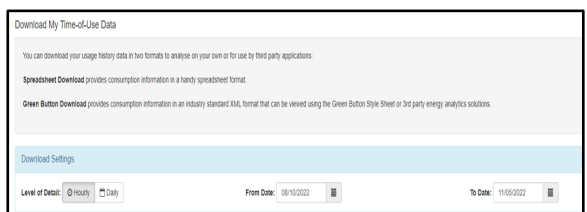

Fig. 6. User data download portal for Waterloo North Hydro

It is worth noting that the portal only permits downloading usage data from the last three months.

The raw dataset lists the power consumption in kWh for each hour on any given day, as seen in Fig. 7.

Fig. 7. Sample raw dataset of user's power consumption

However, this format posed a challenge for the proposed models. Therefore, the data is transposed to meet the appropriate formats for the considered models, as shown in Fig. 8.

Fig. 8. Transposed raw dataset snapshot for model fitting

As seen above, the transposed dataset lists the reading date over 24 hours and the actual power consumption readings. This makes it easier to feed into the proposed models. The dataset is uploaded to a Jupyter notebook, hosted on a Google Cloud instance using TensorFlow/Keras framework, for further processing. During the early stages, a few steps are taken to clean/preprocess the dataset, such as removing entries where the KWh readings are zero resulting in 2090 data rows to be fed to the prediction model. The dataset index was also replaced with the reading date elements, as shown in Fig. 9.

Fig. 9. A part of data preprocessing, cleaning null values

Exploring the dataset shows (Fig. 10) almost a uniform power consumption pattern with random spikes in August with a maximum value of 3.36 Kwh.

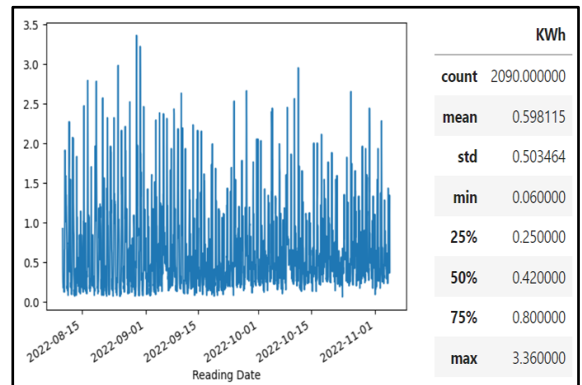

Fig. 10. Exploring the dataset for outliers

Further processing is still needed to prepare the dataset for the proposed models. First, using Sklearn's function "MinMaxScaler", the dataset was normalized to be more suited for the Neural networks-based models as the maximum KWh value is now one. This is to minimize the effect of outliers, as in Fig. 11.

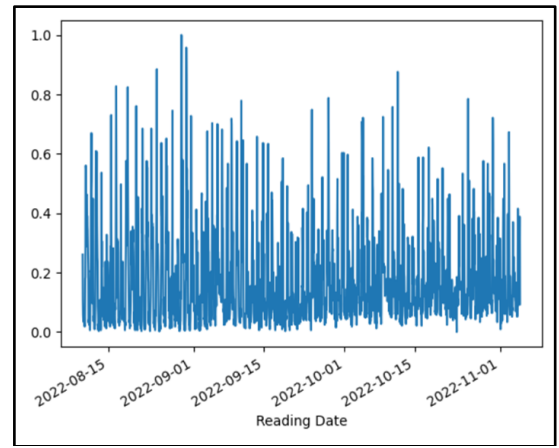

Fig. 11. Daily electricity consumption trend after data normalization

The next preprocessing stage is to transform the dataset into a labeled Matrix by dividing the data elements under the "KWh" column in the dataset into x and y columns utilizing a window size value. The window size is a sliding window that specifies how many KWh readings are used to predict one future value. In our experiment, the model performed better with a window size of 6. It essentially means that the model considers the past 6 actual readings to predict the next value. The matrix, in this case, comprises 6 x elements and a single y element. A function is built to transform the dataset into two arrays, x and y, as seen in the resultant arrays shown in Figures 12 and 13.

```
array([[[0.26060606],
        [0.10909091],
        [0.05757576],
        [0.04242424],
        [0.05454545],
        [0.03333333]],

       [[0.10909091],
        [0.05757576],
        [0.04242424],
        [0.05454545],
        [0.03333333],
        [0.03939394]],

       [[0.05757576],
        [0.04242424],
        [0.05454545],
        [0.03333333],
        [0.03939394],
        [0.05757576]],

       ...,

       [[0.23636364],
        [0.1       ],
        [0.09090909],
        [0.12424242],
        [0.12121212]],
```

Fig. 12. The composed X array

```
array([0.03939394, 0.05757576, 0.03030303, ..., 0.23939394, 0.38787879,
       0.09393939])
```

Fig. 13. The resulting Y array

The new matrix dimensions are reduced to 2084 rows, down from 2090 rows in the dataset. Finally, the y and x arrays are divided into training, validation, and testing sets using 80%, 10%, and 10% percentages, respectively, as shown in Fig. 14.

```
x_train, y_train =x[:1672], y[:1672]
x_val, y_val =x[1672:1881], y[1672:1881]
x_test, y_test =x[1881:], y[1881:]
x_train.shape,y_train.shape, x_val.shape,y_val.shape,x_test.shape,y_test.shape

((1672, 6, 1), (1672,), (209, 6, 1), (209,), (203, 6, 1), (203,))
```

Fig. 14. Snapshot of coding script for training data, validation data, and testing data

### B. Data Analysis and Discussion

As indicated in previous sections, the loss is defined as the Mean Squared Error for both models. Plotting the number of epochs against the loss values for each iteration shows a sharp drop in the loss while the epoch numbers are still very low. It stabilizes around 50 epochs for the LSTM model, indicating that the 100 epochs used in this experiment are overkill. It is worth noting that the GRU model reaches stability faster at around 30 epochs compared to the LSTM mode, as in Figures 15 and 16.

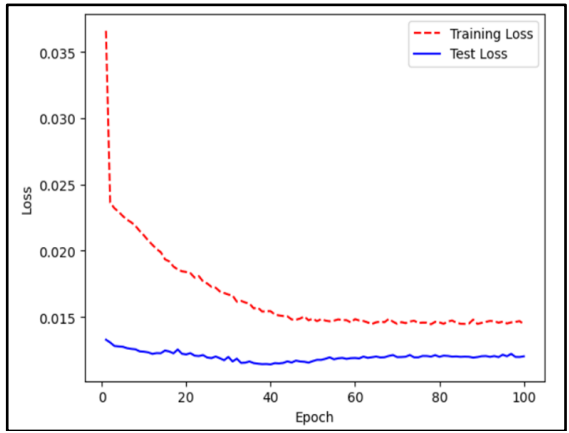

Fig. 15. Loss Vs. the number of epochs for the LSTM model

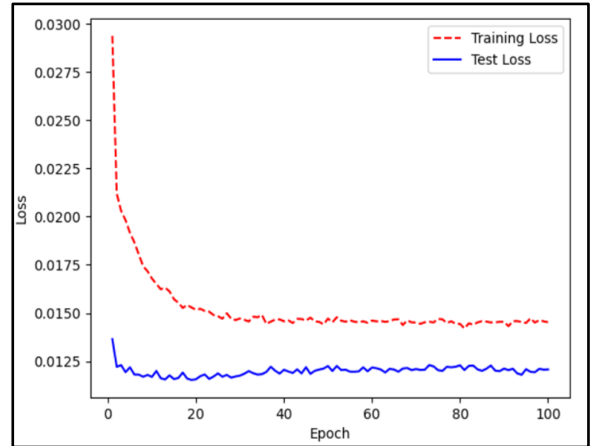

Fig. 16. Loss Vs. the number of epochs for the GRU model

Both models are not overfitting or underfitting, as the values of training loss and validation loss appear close. In both cases, the loss value is very close to zero, indicating a great performance, as shown in Fig. 17.

```
Epoch 98/100
27/27 [==============================] - 1s 27ms/step - loss: 0.0148 - root_mean_squared_error: 0.1216 - val_loss: 0.0120 - val_root_mean_squ
ared_error: 0.1097
```

Fig. 17. Loss value output for overfitting/underfitting testing

Now that the models are well-fitted and display excellent loss readings, the experiment proceeded to the prediction stage using the 'x_train' dataset. However, when the predicted values were placed in a tabular form against the actual values, there was a bit of discrepancy as opposed to what was expected from the models, as shown in Fig. 18.

| | Train predictions | Actuals |
|---|---|---|
| 0 | 0.080202 | 0.039394 |
| 1 | 0.083805 | 0.057576 |
| 2 | 0.094553 | 0.030303 |
| 3 | 0.084051 | 0.024242 |
| 4 | 0.077175 | 0.018182 |
| 5 | 0.072829 | 0.045455 |
| 6 | 0.085851 | 0.178788 |
| 7 | 0.157940 | 0.184848 |
| 8 | 0.180839 | 0.369697 |
| 9 | 0.274774 | 0.560606 |
| 10 | 0.388287 | 0.548485 |
| 11 | 0.398333 | 0.287879 |
| 12 | 0.258509 | 0.457576 |
| 13 | 0.308636 | 0.463636 |
| 14 | 0.328070 | 0.381818 |
| 15 | 0.290708 | 0.324242 |

Fig. 18. Training dataset to calculate prediction values, x_train

However, when plotting the predicted against actual values, the predicated curve follows closely to the actual values curve. The following graph in Fig. 19 displays the predicted values in blue for the last 300 hours.

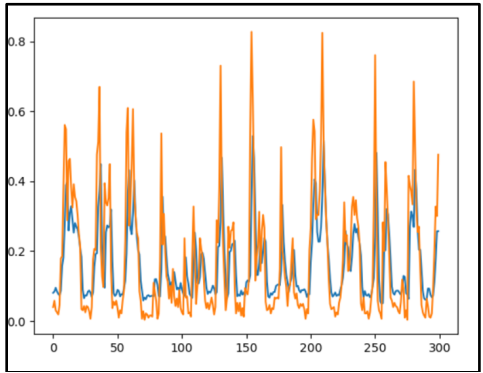

Fig. 19. Actual vs. predicted value plots for the training dataset

Similar trends are observed for predictions based on validation and testing datasets in both models: Based on the validation dataset (for the last 200 hours), indicated in Fig. 20. Based on the testing dataset as shown in Fig. 21 (for the last 200 hours).

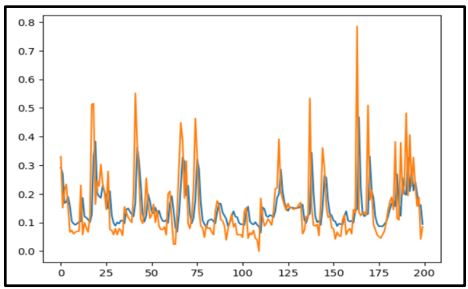

Fig. 20. Actual vs. predicted value plots for validation dataset

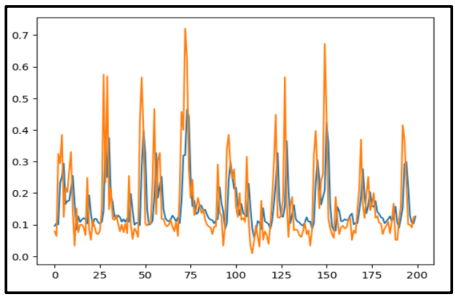

Fig. 21. Actual vs. predicted value plots for the testing dataset

In addition to the RNN-based models, such as LSTM and GRU in our project, we also worked to develop a statistical-based model, ARIMA (Auto Regressive Integrated Moving Average), to perform short-term load forecasting, to learn how the conventional model works in contrast with the AI-based model for a certain type of work. The ARIMA model utilizes the same dataset, but the preprocessing stages are slightly different. For instance, the ARIMA model does not necessarily require the dataset to be normalized; and for this reason, normalization was not utilized in this mode. The first step in building the ARIMA model is to perform a stationarity check, which is done using the following function based on statsmodels (Fig. 22).

```
from statsmodels.tsa.stattools import adfuller
def ad_test(dataset):
    dftest = adfuller(dataset, autolag = 'AIC')
    print("A. ADF : ",dftest[0])
    print("B. P-Value : ", dftest[1])
    print("C. Num Of Lags : ", dftest[2])
    print("D. Num Of Observations Used For ADF Regression:",     dftest[3])
    print("F. Critical Values :")
    for key, val in dftest[4].items():
        print("\t",key, ": ", val)
ad_test(df['KWh'])
```

Fig. 22. Snapshot of coding script for stationarity check in ARIMA

The above functions produce the following results, showing that the data is stationary as the p-value is less than .05.

- ADF:  -6.475502795174342
- P-Value:  1.3343141898787669e-08
- Num Of Lags:  23
- Num of observations used for ADF Regression: 2066.
- Critical Values:
    o 1%:  -3.433519140120394
    o 5%:  -2.862939980034572
    o 10%:  -2.567515285397938

Luckily, a readily available library, pmdarima, can be installed to help determine the order of the AR, I, and MA parts, typically denoted with p, d, and q, respectively, as shown in Fig.23.

```
from pmdarima import auto_arima
stepwise_fit = auto_arima(df['KWh'], trace=True,
suppress_warnings=True)
```

Fig. 23. Snapshot of coding script for ARIMA library import

Based on the following results, the model used in this experiment is of the order (2,0,1), with the lowest AIC value score of 2018.201.

Performing stepwise search to minimize AIC.
- ARIMA(2,0,2)(0,0,0)[0] intercept    : AIC=2036.534, Time=1.60 sec
- ARIMA(0,0,0)(0,0,0)[0] intercept    : AIC=3065.667, Time=0.14 sec
- ARIMA(1,0,0)(0,0,0)[0] intercept    : AIC=2035.007, Time=0.10 sec
- ARIMA(0,0,1)(0,0,0)[0] intercept    : AIC=2315.205, Time=0.19 sec

- ARIMA(0,0,0)(0,0,0)[0]                : AIC=4903.836, Time=0.05 sec
- ARIMA(2,0,0)(0,0,0)[0] intercept    : AIC=2035.977, Time=0.18 sec
- ARIMA(1,0,1)(0,0,0)[0] intercept    : AIC=2036.034, Time=0.26 sec
- ARIMA(2,0,1)(0,0,0)[0] intercept    : AIC=2018.201, Time=1.27 sec
- ARIMA(3,0,1)(0,0,0)[0] intercept    : AIC=2036.627, Time=2.89 sec
- ARIMA(1,0,2)(0,0,0)[0] intercept    : AIC=2037.187, Time=1.12 sec
- ARIMA(3,0,0)(0,0,0)[0] intercept    : AIC=2037.366, Time=0.60 sec
- ARIMA(3,0,2)(0,0,0)[0] intercept    : AIC=2040.812, Time=1.54 sec
- ARIMA(2,0,1)(0,0,0)[0:AIC=inf, Time=0.76 sec
- Best model:  ARIMA (2,0,1)(0,0,0)[0]
- Total fit time: 10.731 seconds

After finding the best ARIMA model, the dataset is split into a training, validation, and testing dataset using the following code, as shown in Fig. 24.

```
train=df.iloc[:1672]
validation=df.iloc[1672:1881]
test=df.iloc[1881:]
print(train.shape,validation.shape,test.shape)
test.head(-25)
```

(1672, 1) (209, 1) (209, 1)

Fig. 24. Training, testing, and validating datasets in ARIMA

Finally, the actual ARIMA model is built, as seen in the screenshot (Fig. 25).

```
from statsmodels.tsa.arima.model import ARIMA
model=ARIMA(train['KWh'],order=(2,0,1))
model=model.fit()
model.summary()
```

Fig. 25. Snapshot of coding script for final ARIMA-based model

Essentially, the model was provided with the training set and the order of the ARIMA model selected at earlier stages. The following in Fig. 26 is the output of the model summary, which lists the coefficients of each AR and MA term. After the model was fitted, the following step in Fig. 27 is making the first prediction using the training data. The first prediction diagram in Fig. 28 shows the predicted values vs the actual training data. Figures 27 and 28 resemble the diagram produced by the LSTM and GRU models. Similar observations are noted when the test set is run through the model, as shown in Fig.29. To evaluate the accuracy of the forecast model, several accuracy metrics were used, such as Mean Squared Error (MSE), Root Mean Squared Error (RMSE), R-squared Score, Mean Absolute error (MAE), and Mean Absolute Percentage Error (MAPE).

**SARIMAX Results**

| Dep. Variable: | KWh | No. Observations: | 1672 |
|---|---|---|---|
| Model: | ARIMA(2, 0, 1) | Log Likelihood | -835.531 |
| Date: | Sun, 04 Dec 2022 | AIC | 1681.061 |
| Time: | 14:53:15 | BIC | 1708.170 |
| Sample: | 08-09-2022 | HQIC | 1691.105 |
|  | - 10-18-2022 | | |
| Covariance Type: | opg | | |

|  | coef | std err | z | P>|z| | [0.025 | 0.975] |
|---|---|---|---|---|---|---|
| const | 0.6016 | 0.042 | 14.197 | 0.000 | 0.519 | 0.685 |
| ar.L1 | 0.7689 | 0.643 | 1.195 | 0.232 | -0.492 | 2.030 |
| ar.L2 | -0.0986 | 0.418 | -0.236 | 0.814 | -0.919 | 0.722 |
| ma.L1 | -0.0958 | 0.646 | -0.148 | 0.882 | -1.361 | 1.170 |
| sigma2 | 0.1590 | 0.004 | 39.959 | 0.000 | 0.151 | 0.167 |

| Ljung-Box (L1) (Q): | 0.00 | Jarque-Bera (JB): | 3094.68 |
|---|---|---|---|
| Prob(Q): | 1.00 | Prob(JB): | 0.00 |
| Heteroskedasticity (H): | 0.83 | Skew: | 1.73 |
| Prob(H) (two-sided): | 0.02 | Kurtosis: | 8.69 |

Fig. 26. A summary of the Output of ARIMA based model

```
start_date='2022-08-09'
end_date='2022-10-17'
pred=model.predict(start=start_date, end=end_date, typ="levels").rename('ARIMA Predictions')
pred
pred.plot(legend=True)
train['KWh'].plot(legend=True, figsize = (16,5))
```

Fig. 27. The first-time prediction in the ARIMA-based model

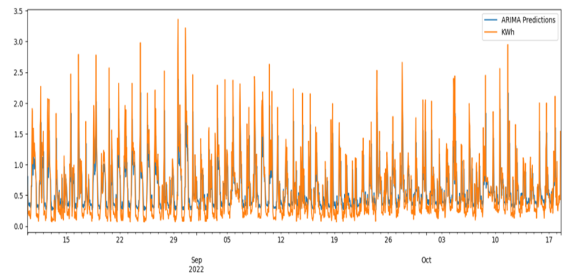

Fig. 28. Actual vs. predicted value plots for first-time prediction in the ARIMA-based model

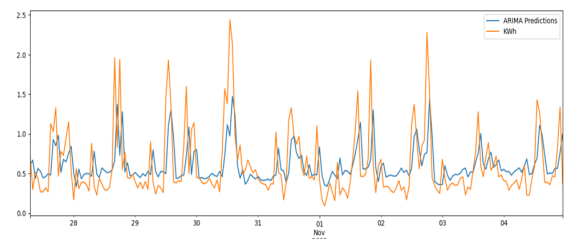

Fig. 29. Actual vs. predicted value plots for testing dataset in the ARIMA-based model

The sklearn-based code (Fig. 30) was utilized to obtain the metrics results. Finally, when all three models (LSTM, GRU, ARIMA) have produced the load forecast prediction, we assessed the obtained results as summarized in Table 2.

```
Arima_r2_score = r2_score(test, pred)
print(Arima_r2_score)
Arima_mse_score = mean_squared_error(test, pred)
print(Arima_mse_score)
Arima_rmse_score = mean_squared_error(test, pred, squared=False)
print(Arima_rmse_score)
Arima_mae_score = mean_absolute_error(test, pred)
print(Arima_mae_score)
Arima_mape_score = mean_absolute_percentage_error(test, pred)
print(Arima_mape_score)
```

Fig. 30. E evaluating the output of the ARIMA-based model

TABLE II. TABLE 2. EVALUATION COMPARISON OF LSTM, GRU, AND ARIMA-BASED MODELS

| Evaluation Method | LSTM model | GRU model | ARIMA model |
|---|---|---|---|
| MSE | 0.0132 | 0.0134 | 0.1330 |
| RMSE | 0.1150 | 0.1157 | 0.3647 |
| R-squared Score | 0.2232 | 0.2054 | 0.2664 |
| MAE | 0.0769 | 0.0791 | 0.2593 |
| MAPE | 0.5642 | 0.5965 | 0.5082 |

LSTM and GRU performed almost identically based on the MSE, RMSE, R-Squared, MAE, and MAPE scores (LSTM performed marginally better). It is worth noting that the GRU-based model can do with fewer epochs compared to the LSTM model, which indicates it is more efficient as it uses less memory and fewer training parameters. Furthermore, although we were methodical in selecting the appropriate ML techniques for their chosen research topic and the dataset type being considered, we do not believe both methods are well suited for small datasets after all. This is evident in the discrepancies between the predicted and the actual values. At the same time, the models still performed very well, as seen in the graphical representation of plotting the predicted and actual values. Both curves follow similar trends. The results also indicate that the discrepancies are not attributed to the model's over or underfitting. The loss values across all datasets (actual, validation, and testing) are very close. The statistical-based model, ARIMA model in our case, performed less accurately compared to LSTM and GRU, which is expected as ARIMA is not ideal for short-term forecasting exercises. Most of the parameters applied to the models were based on long trials and errors, which aid in fine-tuning the models and producing favorable results. However, we believe fewer numbers of epochs could have produced similar results. The models performed poorly, with higher window sizes for more than 24 hours. Both Tanh and ReLU activation functions provided similar results.

## VI. CONCLUSION AND FUTURE DISCUSSION

This paper focused on building a forecast model suitable for predicting short-term power consumption at a household level using customized deep-learning models such as LSTM and GRU. Both models are still ideal for time series datasets, but their main advantage comes in handy when creating a long-term forecasting model. Acquiring a larger dataset and using long-term forecasting models such as SARIMA should be investigated as a future extension of this paper. Future directions include the application of ensemble learning [39]-[42], multi-level learning [43][44], large-scale optimizations [45]-[46], and simulation models [47].

## REFERENCES

[1] S. H. Rafi, Nahid-Al-Masood, S. R. Deeba, and E. Hossain, "A Short-Term Load Forecasting Method Using Integrated CNN and LSTM Network," in IEEE Access, vol. 9, pp. 32436-32448, 2021, DOI: 10.1109/ACCESS. 2021.3060654.

[2] M. Jacob, C. Neves, and D. Vukadinovi´c Greetham, "Introduction," in Forecasting and Assessing Risk of Individual Electricity Peaks, M. Jacob, C. Neves, and D. Vukadinovi´c Greetham, Eds. Cham: Springer International Publishing, 2020, pp. 1–14.

[3] W. Kong, Z. Y. Dong, Y. Jia, D. J. Hill, Y. Xu, and Y. Zhang, "Short-Term Residential Load Forecasting Based on LSTM Recurrent Neural Network," in IEEE Transactions on Smart Grid, vol. 10, no. 1, pp. 841-851, Jan. 2019, DOI: 10.1109/TSG.2017.2753802.

[4] X. Guo, Q. Zhao, S. Wang, D. Shan, and W. Gong, "A short-term load forecasting model of LSTM neural network considering demand response," Complexity, vol. 2021, 2021.

[5] D. Syed, H. Abu-Rub, A. Ghrayeb, and S. S. Refaat, "Household-Level Energy Forecasting in Smart Buildings Using a Novel Hybrid Deep Learning Model," in IEEE Access, vol. 9, pp. 33498-33511, 2021, DOI: 10.1109/ACCESS.2021.3061370.

[6] Nourhan M. Ibrahim, Ashraf I. Megahed, Nabil H. Abbasy, "Short-Term Individual Household Load Forecasting Framework Using LSTM Deep Learning Approach," 5th International Symposium of Multidisciplinary Studies and Innovative Technologies (ISMSIT), 2021

[7] A. Gs and K. Shiruru, "Short Term Load Forecasting Methods, A Comparative Study," International Journal of Advance Research and Innovative Ideas in Education, vol. 1, 2016.

[8] H.-X. Zhao and F. Magoulès, "A review on the prediction of building energy consumption," Renew. Sustain. Energy Rev., vol. 16, no. 6, pp. 3586–3592, 2012.

[9] M. A. Hammad, B. Jereb, B. Rosi, and D. Dragan, "Methods and models for electric load forecasting: A comprehensive review," Logist. sustain. transp., vol. 11, no. 1, 2020.

[10] A. Rahman, V. Srikumar, and A. D. Smith, "Predicting electricity consumption for commercial and residential buildings using deep recurrent neural networks," Appl. Energy, vol. 212, 2018.

[11] D. Syed, H. Abu-Rub, A. Ghrayeb, and S. S. Refaat, "Household-Level Energy Forecasting in Smart Buildings Using a Novel Hybrid Deep Learning Model," in IEEE Access, vol. 9, pp. 33498-33511, 2021, DOI: 10.1109/ACCESS.2021.3061370.

[12] Y. Bengio, A. Courville, and P. Vincent, "Representation learning: A review and new perspectives," IEEE Trans. Pattern Anal. Mach. Intell., vol. 35, no. 8, pp. 1798–1828, Aug. 2013.

[13] S. Hochreiter and J. Schmidhuber, "Long short-term memory," Neural Comput., vol. 9, no. 8, pp. 1735–1780, 1997.

[14] I. Sutskever, O. Vinyals, and Q. V. Le, "Sequence to sequence learning with neural networks," in Proc. Adv. Neural Inf. Process. Syst., Montreal, QC, Canada, 2014, pp. 3104–3112.

[15] Ibrahim, A., Kashef, R., & Corrigan, L. (2021). Predicting market movement direction for bitcoin: A comparison of time series modeling methods. Computers & Electrical Engineering, 89, 106905.

[16] Ibrahim, A., Kashef, R., Li, M., Valencia, E., & Huang, E. (2020). Bitcoin network mechanics: Forecasting the btc closing price using vector auto-regression models based on endogenous and exogenous feature variables. Journal of Risk and Financial Management, 13(9), 189.

[17] Tan, X., & Kashef, R. (2019, December). Predicting the closing price of cryptocurrencies: a comparative study. In Proceedings of the Second International Conference on Data Science, E-Learning and Information Systems (pp. 1-5).

[18] Tobin, T., & Kashef, R. (2020, June). Efficient Prediction of Gold Prices Using Hybrid Deep Learning. In International Conference on Image Analysis and Recognition (pp. 118-129). Springer, Cham.

[19] X. Cao, S. Dong, Z. Wu, and Y. Jing, "A data-driven hybrid optimization model for short-term residential load forecasting," in Proc. IEEE Int. Conf. Comput. Inf. Technol. Ubiquitous Comput. Commun. Dependable Auton. Secure Comput. Pervasive Intell. Comput. (CIT/IUCC/DASC/PICOM), K., 2015, pp. 283–287.

[20] Z. Yun et al., "RBF neural network and ANFIS-based short-term load forecasting approach in real-time price environment," IEEE Trans. Power Syst., vol. 23, no. 3, pp. 853–858, Aug. 2008.

[21] H. Li, Y. Zhao, Z. Zhang, and X. Hu, "Short-term load forecasting based on the grid method and the time series fuzzy load forecasting method," in Proc. Int. Conf. Renew. Power Gener. (RPG), 2015.

[22] P. Qingle and Z. Min, "Very short-term load forecasting based on neural network and rough set," in Proc. Int. Conf. Intell. Comput. Technol. Autom. (ICICTA), China, 2010, pp. 1132–1135.

[23] R. Zhang, Z. Y. Dong, Y. Xu, K. Meng, and K. P. Wong, "Short-term load forecasting of Australian national electricity market by an ensemble model of extreme learning machine," IET Gener. Transm. Distrib., vol. 7, no. 4, pp. 391–397, Apr. 2013.

[24] R. Zhang, Y. Xu, Z. Y. Dong, W. Kong, and K. P. Wong, "A composite k-nearest neighbor model for day-ahead load forecasting with limited temperature forecasts," presented at the IEEE Gen. Meeting, Boston, MA, USA, 2016, pp. 1–5.

[25] F. H. Al-Qahtani and S. F. Crone, "Multivariate k-nearest neighbour regression for time series data—A novel algorithm for forecasting UK electricity demand," in Proc. Int. Joint Conf. Neural Netw. (IJCNN), Dallas, TX, USA, 2013, pp. 1–8.

[26] M. Ghofrani, M. Ghayekhloo, A. Arabali, and A. Ghayekhloo, "A hybrid short-term load forecasting with a new input selection framework," Energy, vol. 81, pp. 777–786, Mar. 2015.

[27] P. Zhang, X. Wu, X. Wang, and S. Bi, "Short-term load forecasting based on big data technologies," CSEE J. Power Energy Syst., vol. 1, no. 3, pp. 59–67, Sep. 2015.

[28] F. L. Quilumba, W.-J. Lee, H. Huang, D. Y. Wang, and R. L. Szabados, "Using smart meter data to improve the accuracy of intraday load forecasting considering customer behavior similarities," IEEE Trans. Smart Grid, vol. 6, no. 2, 2015.

[29] B. Stephen, X. Tang, P. R. Harvey, S. Galloway, and K. I. Jennett, "Incorporating practice theory in sub-profile models for short term aggregated residential load forecasting," IEEE Trans. Smart Grid, vol. 8, no. 4, pp. 1591–1598, Jul. 2017.

[30] M. Chaouch, "Clustering-based improvement of nonparametric functional time series forecasting: Application to intra-day household-level load curves," IEEE Trans. Smart Grid, vol. 5, no. 1, pp. 411–419, Jan. 2014.

[31] M. Ghofrani, M. Hassanzadeh, M. Etezadi-Amoli, and M. S. Fadali, "Smart meter based short-term load forecasting for residential customers," in Proc. North Amer. Power Symp. (NAPS), Boston, MA, USA, 2011, pp. 1–5.

[32] S. Ryu, J. Noh, and H. Kim, "Deep neural network based demand side short term load forecasting," in Proc. IEEE Int. Conf. Smart Grid Commun. (SmartGridComm), Sydney, NSW, Australia, 2016, pp. 308–313.

[33] E. Mocanu, P. H. Nguyen, M. Gibescu, and W. L. Kling, "Deep learning for estimating building energy consumption," Sustain. Energy Grids Netw., vol. 6, pp. 91–99, Jun. 2016.

[34] D. L. Marino, K. Amarasinghe, and M. Manic, "Building energy load forecasting using deep neural networks," in Proc. 42nd Annu. Conf. IEEE Ind. Electron. Soc. (IECON), Florence, Italy, 2016, pp. 7046–7051.

[35] Y. Bengio, P. Simard and P. Frasconi, "Learning long-term dependencies with gradient descent is difficult," in IEEE Transactions on Neural Networks, vol. 5, no. 2, 1994.

[36] Cristina Heghedus, Antorweep Chakravorty, Chunming Rong, "Energy Load Forecasting Using Deep Learning," 2018 IEEE International Conference on Energy Internet (ICEI), 2018

[37] K. Cho, Learning Phrase Representations using RNN Encoder-Decoder for Statistical Machine Translation, arXiv:1406.1078, 2014.

[38] Junyoung Chung, Caglar Gulcehre, KyungHyun Cho, Yoshua Bengio, Empirical Evaluation of Gated Recurrent Neural Networks on Sequence Modeling, arXiv:1412.3555v1, [cs.NE] 11 Dec 2014

[39] Kashef, R. F. (2018, January). Ensemble-based anomaly detetction using cooperative learning. In KDD 2017 Workshop on Anomaly Detection in Finance (pp. 43-55). PMLR.

[40] Soleymanzadeh, R., Aljasim, M., Qadeer, M. W., & Kashef, R. (2022). Cyberattack and Fraud Detection Using Ensemble Stacking. AI, 3(1), 22-36.

[41] Aljasim, M., & Kashef, R. (2022). E2DR: a deep learning ensemble-based driver distraction detection with recommendations model. Sensors, 22(5), 1858.

[42] Ebrahimian, M., & Kashef, R. (2021, September). A CNN-based Hybrid Model and Architecture for Shilling Attack Detection. In 2021 IEEE Canadian Conference on Electrical and Computer Engineering (CCECE) (pp. 1-7). IEEE.

[43] Aasi, B., Imtiaz, S. A., Qadeer, H. A., Singarajah, M., & Kashef, R. (2021, April). Stock Price Prediction Using a Multivariate Multistep LSTM: A Sentiment and Public Engagement Analysis Model. In 2021 IEEE International IOT, Electronics and Mechatronics Conference (IEMTRONICS) (pp. 1-8). IEEE.

[44] Li, M., Kashef, R., & Ibrahim, A. (2020). Multi-level clustering-based outlier's detection (MCOD) using self-organizing maps. Big Data and Cognitive Computing, 4(4), 24.

[45] Kashef, R., & Niranjan, A. (2017, December). Handling Large-Scale Data Using Two-Tier Hierarchical Super-Peer P2P Network. In Proceedings of the International Conference on Big Data and Internet of Thing (pp. 52-56).

[46] Manjunath, Y. S. K., & Kashef, R. F. (2021). Distributed clustering using multi-tier hierarchical overlay super-peer peer-to-peer network architecture for efficient customer segmentation. Electronic Commerce Research and Applications, 47, 101040.

[47] A. Aldhubaib and R. Kashef, "Optimizing the Utilization Rate for Electric Power Generation Systems: A Discrete-Event Simulation Model," in IEEE Access, vol. 8, pp. 82078-82084, 2020, doi: 10.1109/ACCESS.2020.2991362.